\documentclass[11pt]{article}

\newsavebox{\junk}
\savebox{\junk}[1.6mm]{\hbox{$|\!|\!|$}}

\def\bbbz{{\mathchoice {\hbox{$\sf\textstyle Z\kern-0.4em Z$}}
{\hbox{$\sf\textstyle Z\kern-0.4em Z$}}
{\hbox{$\sf\scriptstyle Z\kern-0.3em Z$}}
{\hbox{$\sf\scriptscriptstyle Z\kern-0.2em Z$}}}}
\def\sq{\hbox{\rlap{$\sqcap$}$\sqcup$}}

\usepackage{epsfig} 
\usepackage{amsfonts}
\usepackage{latexsym}
\usepackage{bm}

\usepackage{pstricks}
\usepackage{colortab}
\usepackage{tikz}

\definecolor{lightgray}{rgb}{0.75,0.75,0.75}

\newcommand{\ben}{\begin{enumerate}}
\newcommand{\een}{\end{enumerate}}
\newcommand{\bit}{\begin{itemize}}
\newcommand{\eit}{\end{itemize}}

\newtheorem{theorem}{Theorem}[section]
\newtheorem{proposition}[theorem]{Proposition}

\newcommand{\ba}{\begin{array}{rcl}}
\newcommand{\ea}{\end{array}}
\newcommand{\bt}{\begin{theorem}}
\newcommand{\et}{\end{theorem}}
\newcommand{\bd}{\begin{description}}
\newcommand{\ed}{\end{description}}

\def\slabel#1{\label{s:#1}}

\def\elabel#1{\label{e:#1}}
\def\eq#1/{(\ref{e:#1})}

\def\Section#1{Section~\ref{s:#1}}

\def\beq{\begin{equation}}
\def\eeq{\end{equation}}
\def\beqa{\begin{eqnarray}}
\def\eeqa{\end{eqnarray}}

\def\qed{\ifmmode\sq\else{\unskip\nobreak\hfil
\penalty50\hskip1em\null\nobreak\hfil\sq
\parfillskip=0pt\finalhyphendemerits=0\endgraf}\fi}

\def\sqr#1#2{{\vcenter{\hrule height.#2pt
      \hbox{\vrule width.#2pt height#1pt \kern#1pt
         \vrule width.#2pt}
       \hrule height.#2pt}}}

\newcommand{\bc}{\begin{corollary}}
\newcommand{\ec}{\end{corollary}}

\newcommand{\bp}{\begin{proposition}}
\newcommand{\ep}{\end{proposition}}

\def\eye(#1){{\bf (#1)}\quad}

\def\taboo#1{{{}_{#1}}}

\def\0P{\taboo{0}P}
\def\0Pn{\taboo{0}P^n}

\def\divline{\begin{center} \line(1,0){200} \end{center}}

\title{An Efficient Search Strategy for Aggregation and Discretization of Attributes of Bayesian Networks Using Minimum Description Length}

\usepackage{authblk}
\usepackage[hang,flushmargin]{footmisc}

\author[1]{J.N. Corcoran}
\author[2]{D. Tran}
\author[3]{N.D. Levine}
\affil[1]{University of Colorado, Boulder}
\affil[2]{University of Kansas}
\affil[3]{Naval Postgraduate School}

\begin{document}


\maketitle
\vspace{-.8cm}

\begin{abstract} \small \noindent
Bayesian networks are convenient graphical expressions 
for  high dimensional probability distributions representing complex 
relationships between a large number of random variables. They have been employed extensively in areas such as bioinformatics, artificial intelligence, diagnosis, and risk management. The recovery of the structure of a network from data is of prime importance for the purposes of modeling, analysis, and prediction. Most recovery algorithms in the literature assume either discrete of continuous but Gaussian data. For general continuous data, discretization is usually employed but often destroys the very structure one is out to recover. Friedman and Goldszmidt \cite{friedgold96} suggest an approach based on the minimum description length principle that chooses a discretization which preserves the information in the original data set, however it is one which is difficult, if not impossible, to implement for even moderately sized networks. In this paper we provide an extremely efficient search strategy which allows one to use the Friedman and Goldszmidt discretization in practice.

\bigskip

\end{abstract}
\let\thefootnote\relax\footnotetext{$^{*}$Postal Address:
      J.N. Corcoran,
      Department of Applied Mathematics,
      University of Colorado, Box 526
      Boulder CO 80309-0526, USA; email: corcoran@colorado.edu; 
      phone: 303-492-0685}
\let\thefootnote\relax\footnotetext{Keywords: Bayesian networks, discretization, minimum description length \\
AMS Subject classification: 62-09, 68P30, 62C99}

\setcounter{page}{0}



\section{Introduction}
\slabel{int} 
Bayesian networks are convenient graphical expressions for  high dimensional 
probability distributions representing complex relationships between a large 
number of random variables. They have been employed extensively in areas such 
as bioinformatics (\cite{bockhorst2003}, \cite{friedman00}, \cite{mourad2011},
 \cite{larranaga2005}), artificial intelligence (\cite{korb2003}, 
\cite{griffiths}, \cite{tenenbaum}), diagnosis (\cite{curiac}, \cite{onisko}, \cite{xiang}), and risk management (\cite{cowell}, \cite{fenton2013}, \cite{weber}) for the purposes of general modeling, prediction, and diagnosis.

\begin{figure}[htp]
\caption{A Directed Acyclic Graph}
\label{fig:onedag}
\begin{center}
\ \psfig{figure=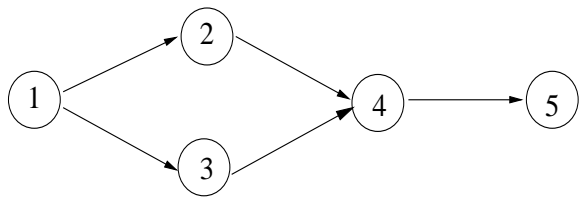,height=1in,width=3in} \\
\end{center}
\end{figure}

A Bayesian network consists of a directed acyclic graph (DAG), in which nodes represent random variables,  and a set of conditional probability distributions. For example, 
the graph in Figure \ref{fig:onedag} represents a network for 5 variables, $X_{1}$, $X_{2}$, $X_{3}$, $X_{4}$, and $X_{5}$ with a joint probability density $p(x_{1}, x_{2}, x_{3}, x_{4}, x_{5})$ which is assumed to factor as
$$
p(x_{1}, x_{2}, x_{3}, x_{4}, x_{5}) = p(x_{1}) \cdot p(x_{2}|x_{1}) \cdot p(x_{3}|x_{1}) \cdot p(x_{4}|x_{2},x_{3}) \cdot p(x_{5}|x_{4}).
$$

In general, for a Bayesian network on $n$ nodes, the joint density for the corresponding $n$ random variables may be written as
\beq 
\elabel{factor}
p(x_{1}, x_{2}, \ldots, x_{n}) = \prod_{i=1}^{n} p(x_{i}|\Pi_{i})
\eeq
where $\Pi_{i}$ is used to denote the set of random variables corresponding to ``parent'' nodes for node $i$. For the example in Figure \ref{fig:onedag}, we have $\Pi_{1} = \emptyset$, $\Pi_{2} = \{x_{1}\}$, $\Pi_{3}=\{x_{1}\}$, $\Pi_{4} = \{x_{2},x_{3}\}$, and $\Pi_{5} = \{x_{4}\}$. 

Given $m$ observations of $(X_{1}, X_{2}, \ldots, X_{n})$, our ultimate goal is to infer the connecting arrows for a corresponding $n$-node graph. There are many ways to do this rather successfully for discrete data or for Gaussian data. In the case of non-Gaussian continuous data, discretization is usually employed, however care must be taken not to destroy the very structure one wishes to retrieve. Since the problem involves mapping entire intervals of values to discrete values, it is closely related to the ``aggregation problem'' where one reduces the number of possible values taken on by a node by mapping collections of values in already discrete data to single points. (We will think of this as a ``discretization of discrete data'' and will still refer to it as discretization.) 

\begin{figure}[htp]
\caption{A Three Node DAG}
\label{fig:dag8}
\begin{center}
\ \psfig{figure=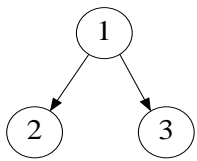,height=0.8in,width=1in} \\
\end{center}
\end{figure}

For an example of a discretization masking important attributes of the underlying network structure, consider the three node DAG show in Figure \ref{fig:dag8} where the corresponding random variables $X_{1}$, $X_{2}$, and $X_{3}$ each take on values in $\{1,2,3\}$.
Consider the discretization to random variables $X_{1}^{*}$, $X_{2}^{*}$, and $X_{3}^{*}$ defined as $X_{i}^{*} = 1$ if $X_{i} \in \{1,2\}$ and  $X_{i}^{*} = 2$ if $X_{i} = 3$.

The original network structure implies that the variables represented by nodes $2$ and $3$ are conditionally independent given the value of the variable at node 1, yet it is easy to verify that we have, after discretization,
$$
P(X_{2}^{*}=1,X_{3}^{*}=1|X_{1}^{*}=1) \ne P(X_{2}^{*}=1|X_{1}^{*}=1) \cdot P(X_{3}^{*}=1|X_{1}^{*}=1),
$$ 
for example. Thus, there is little hope of recovering the correct network structure after this discretization with data that have been discretized in this way.

Friedman and Goldszmidt \cite{friedgold96} suggest an approach based on the {\emph{minimum description length principle}} that chooses a discretization which preserves the information in the original data set. However, their approach, which we describe in \Section{mdl}, requires that we compute and compare scores for every different possible discretization of the data. The number of possibilities is often prohibitively large.

In this paper we give a search strategy in which only a very small subset of minimum description length scores must be computed. Under certain simplifying assumptions, we prove that the procedure will identify the optimal score. Furthermore, in the case where there is a ``correct'' discretization in the sense that distributions of nodes are unchanged over entire intervals of values for children, we show that the optimal scoring network does in fact correspond to the data generating network.

\section{Likelihood-Based Recovery of Bayesian Networks for Discrete Data}
\slabel{bnnotation}

We will use this section to establish some notation that will be used throughout this paper and to briefly review commonly used likelihood-based methods for recovering the structure of Bayesian networks.

Consider a Bayesian network on $n$ nodes. For $i=1,2, \ldots, n$, let $X_{i}$ denote the random variable associated with node $i$. We will refer to node $i$ and the associated random variable $X_{i}$, interchangeably. Suppose that $X_{i}$ can take on values in $\{1,2, \ldots, ||X_{i}||\}$. Let $\Pi_{i}$ denote the set of parent node indices for node $i$, let $|\Pi_{i}|$ denote the number of parents for node $i$, and let $||\Pi_{i}||$ denote the number of possible parent configurations. For example, for the network in in Figure \ref{fig:onedag}, 
$$
|\Pi_{4}| = 2 \qquad \mbox{and} \qquad ||\Pi_{4}|| = ||X_{2}|| \cdot ||X_{3}||.
$$

For $i=1,2, \ldots, n$, $j=1,2, \ldots, ||\Pi_{i}||$, and $k=1,2, \ldots, ||X_{i}||$, define parameters
$$
\theta_{ijk} := P(X_{i}=k|\Pi_{i} = j).
$$
Here, ``$\Pi_{i}=j$'' refers to a specific set of values for the parent variables, configurations of which have been enumerated and are now indexed by $j$.

Now, from the joint density (\ref{e:factor}), the likelihood for the entire $m \times n$ dimensional data set may be written as
$$
L(\theta) := \prod_{i,j,k} \theta_{ijk}^{n_{ijk}}
$$
where $n_{ijk}$ is the total number of times in the sample that $X_{i}$ is observed to have value $k$ when its parents take on configuration $j$. (To keep this concise expression for the likelihood, we define, for a parentless node $i$, $|\Pi_{i}|=1$ and $\theta_{i1k} = P(X_{i}=k)$.)

There are many ways to recover networks from data. Indeed, we may not even want to think in terms of ``one best network'' and instead use a model averaging approach or one that constructs a best network by combining best ``features'' (for example edges) from several networks. In this paper, we will restrict our attention to  simple methods for recovering a single ``best'' network as measured by various standard likelihood and information criterion indices. We will assume a manageable number of networks to score. Of course, the number of possible networks increases superexponentially in the number of nodes-- the results of this paper might then be applied using Monte Carlo search strategies.

\begin{enumerate}
\item Log-Likelihood (LL)

Given $m$ $n$-tuples of data points, we compute the log-likelihood for every possible graph. For each DAG, we have a different set of relevant $\theta$ parameters. Given a particular DAG, we estimate each $\theta$ with its maximum likelihood estimator
$$
\hat{\theta}_{ijk} = \frac{\mbox{\# observations with $X_{i}=k$ and $\Pi_{i}=j$}}{\mbox{\# observations with $\Pi_{i}=j$}},
$$ 
and then we report the log-likelihood
$$
\ln L(\hat{\theta}) = \sum_{i,j,k} n_{ijk} \ln (\hat{\theta}_{ijk}).
$$

As we can always increase the log-likelihood by including additional $\theta$ parameters, we expect the greatest log-likelihoods (``most likely models'') to coincide with DAGs with a maximal number of edges. Therefore, it is important include a penalty for overparameterized models. The two most common penalized log-likelihood statistics are given by the following information criteria.

\item Akaike's Information Criterion (AIC)

Akaike's Information Criterion (AIC) is, in its most general form, defined by
$$
AIC = -2 \ln L + 2 \cdot (\# parameters).
$$ 
Clearly, the goal is to minimize the AIC to ensure a good fitting model in the sense of maximizing the log-likelihood while penalizing for having too many parameters.

\item Bayesian Information Criterion (BIC)

The Bayesian Information Criterion (BIC) is defined by
$$
BIC = -2 \ln L + (\# parameters) \cdot \ln (m),
$$
where $m$ is, as before, the sample size. As with the AIC, the goal is to minimize the BIC. 

\end{enumerate}

Both the AIC and BIC have rigorous justifications, from both Bayesian and frequentist points of view. AIC was derived from information theory though it can be though of as Bayesian if one uses a clever choice of prior. On the other hand BIC, originally derived through Bayesian statistics as a measure of the Bayes factor, can also be derived as a non-Bayesian result. For more information on these widely used scoring criteria, we refer the interested reader to \cite{akaike73},\cite{akaike74}, \cite{akaike81}, and \cite{bozdogan} (AIC), and  \cite{kashvap} and \cite{schwartz} (BIC). To make some broad generalizations, AIC, it can often overfit the model in terms of number of parameters. BIC, on the other hand,  tends to overpenalize, or underfit the model.That being said, we have been most successful in recovering the correct structure of simulated Bayesian networks using AIC. Incidentally, the use of Bayesian networks does not necessarily imply a Bayesian approach to modeling where parameters are considered as random variables. In this paper,  we are taking the frequentist approach.

\section{Minimum Description Length}
\slabel{mdl}

Arguably, the point of statistical modeling is to find regularities in an observed data set. Discovered regularities allow the modeler to be able to describe the data more succinctly. The minimum description length (MDL) principle, introduced in 1978 by Jorma Rissanen \cite{rissanen78}, is a model selection technique that chooses, as the best model, the one that permits the shortest encoding of both the model and the observed data. 

In this section, we describe how Friedman and Goldszmidt \cite{friedgold96} define a description length that can be used for network recovery from  discrete data. It is the approximate length of storage space, measured in bits, for the binary representation of the DAG (network structure), the parameters (the $\theta_{ijk}$) that, together with the DAG, define a Bayesian network, and the data itself.
We will see, in the end, that it is simply another penalized likelihood approach.

In what follows, we repeatedly use the fact that an integer $k$ can be encoded in approximately $\lceil \log_{2} k \rceil$ bits. For example, the decimal value $9$ becomes, in binary, the $\lceil \log_{2} 9 \rceil = 4$ bit number 1001. Throughout this paper, we will use $\log$ to denote the base $2$ logarithm, though, in the end the base is unimportant for the comparisons we will make.

\subsection*{Encoding the DAG}
As in \Section{bnnotation}, we assume that $X_{i}$, can take on $||X_{i}||$ possible values, and that they are integers ranging from $1$ to $||X_{i}||$. (For example, if $X_{1}$ can take on values in $\{2,5,11\}$, we would relabel the values as $1$, $2$, and $3$.)

The approximate number of bits needed to encode the number of nodes/variables $n$, and the number of possible values taken on by each of those variables is
\beq
\elabel{varsandvals}
\log n + \sum_{i=1}^{n} \log ||X_{i}||.
\eeq

In order to completely describe the network structure, we must also include the number of parents for each node and the actual list of parents for each node. 
As a simplification, since the number of parents of node $i$, which is denote by $|\Pi_{i}|$, is always less than $n$, we can, conservatively, reserve $\log n$ bits to encode $|\Pi_{i}|$. Doing this for each node, we add
\beq
\elabel{numberpa}
\sum_{i=1}^{n} \log n = n \log n
\eeq
to our network description length.

For the actual list of parents, since the maximum value in the list of indices is $n$, we will use the conservative value of $\log n$ bits to encode each index. For node $i$, we must encode $|\Pi_{i}|$ different indices, each using a length of $\log n$ bits of space. So, in total, to encode all parent lists, we will use
\beq 
\elabel{parlist}
\sum_{i=1}^{n} |\Pi_{i}| \cdot \log n
\eeq 
bits.

In total, we use
\beq 
\elabel{dagstore}
\log n + \sum_{i=1}^{n} \log ||X_{i}|| +  \sum_{i=1}^{n} (1+|\Pi_{i}|) \cdot \log n
\eeq 
bits to encode the DAG structure.

\subsection*{Encoding the Parameters}
The Bayesian network consists of a DAG together with a collection of parameters $\theta_{ijk} = P(X_{i}=k|\Pi_{i}=j)$. Since $\sum_{k} \theta_{ijk}=1$, we only need to encode $||\Pi_{i}|| \cdot (||X_{i}||-1)$ parameters for node $i$. However, the parameters are not integer valued. In our case, for network recovery, we will actually be storing/encoding parameters that have been estimated from our $m$ $n$-dimensional data points. Friedman and Goldszmidt \cite{friedgold96} indicate that the ``usual choice in the literature'' is to use $\frac{1}{2} \log m$ bits per parameter. Thus, we will use
\beq
\elabel{encodeparams}
\frac{1}{2} \log m \cdot \sum_{i=1}^{n} ||\Pi_{i}|| \cdot (||X_{i}||-1)
\eeq
bits to encode the estimated network parameters.

\vspace{0.1in}
We add (\ref{e:dagstore}) and (\ref{e:encodeparams}) in order to define description length for the Bayesian network. Since we are trying to infer the best connecting arrows for a graph on $n$ nodes, we presumably already know the number of nodes and would not be encoding it. Thus, we drop the $\lceil \log n \rceil$ term and define the {\emph{network description length as}}
\beq
\elabel{DLnet}
DL_{net} = \sum_{i=1}^{n} \log ||X_{i}|| +  \sum_{i=1}^{n} (1+|\Pi_{i}|) \cdot \log n + \frac{1}{2} \log m \cdot \sum_{i=1}^{n} ||\Pi_{i}|| \cdot (||X_{i}||-1).
\eeq

\subsection*{Encoding the Data}

Consider the string of digits $2213$ stored, in binary as $1010111$. (The binary representations of $1$, $2$, and $3$ are $1$, $10$, and $11$, respectively.) Without separators, the binary string $1010111$ can not be decoded near the end. The $111$ maybe be, in decimal, three $1$'s, a $1$ followed by a $3$, or a $3$ followed by a $1$. ``Prefix codes'' avoid this problem by encoding the digits with zeros and ones in a way so that no encoded digit is a prefix of any other encoded digit. Further consideration can be made to ensure the code is not only a prefix code, but one that results in the maximum compression of the data by assigning, to the original decimal digits, binary codes of lengths inversely proportional to the frequencies with which they appear in the original data. 

In the context of Bayesian networks, we wish to encode the values in the $(m \times n)$-dimensional data set with code lengths that are inversely proportional to the frequencies (equivalently, estimated probabilities) with which they appear in the data. Friedman and Goldszmidt \cite{friedgold96} use {\emph{Shannon coding}} \cite{coverthomas2006} which is not optimal in terms of compression, but which encodes each $n$-dimensional data point $\vec{x}_{i}=(x_{i1}, x_{i2}, \ldots, x_{in})$ using approximately $-\log p(\vec{x}_{i})$ bits where $p(\vec{x}_{i})$ is the joint density for $X_{1}, X_{2}, \ldots, X_{n}$ evaluated at $\vec{x}_{i}$. Thus, the entire data set, consisting of $m$ such vectors is encoded in approximately
\beq
\elabel{DLdata}
DL_{data}=-\sum_{i=1}^{m} \log p(\vec{x}_{i})
\eeq
bits. This is desirable from a modeling standpoint since it corresponds to the familiar log-likelihood commonly used in statistical inference.

\divline
Summing (\ref{e:DLnet}) and (\ref{e:DLdata}), we define the description length for a Bayesian network and the observed data as
\beq
\elabel{dl}
DL = \sum_{i=1}^{n} \log ||X_{i}|| +  \sum_{i=1}^{n} (1+|\Pi_{i}|) \cdot \log n + \frac{1}{2} \log m \cdot \sum_{i=1}^{n} ||\Pi_{i}|| \cdot (||X_{i}||-1) -\sum_{i=1}^{m} \log p(\vec{x}_{i}). 
\eeq
Given discrete data and a collection of possible Bayesian networks, the network chosen by the MDL principle is the one that minimizes (\ref{e:dl}). In light of the form of $DL_{data}$, we see that this is simply a penalized log-likelihood scoring metric, similar to the AIC and BIC discussed in \Section{bnnotation}. We now illustrate the performance of all three for a three node network. The list of all 25 DAGs corresponding to 3 node networks can be found in Table \ref{t:3nodeDAGs} and we will refer to these DAGs as they are numbered here.

\begin{table}[ht]
\caption{Directed Acyclic Graphs on Three Nodes}
\label{t:3nodeDAGs}
\begin{center}
\begin{tabular}{|c|c|c|c|c|}  \hline \hline
\begin{minipage}[t]{0.15\textwidth}
\begin{flushleft}\hspace{-0.075in}\fbox{1}\end{flushleft}
\end{minipage}
 &
\begin{minipage}[t]{0.15\textwidth}
\begin{flushleft}\hspace{-0.075in}\fbox{2}\end{flushleft}
\end{minipage}
 &
\begin{minipage}[t]{0.15\textwidth}
\begin{flushleft}\hspace{-0.075in}\fbox{3}\end{flushleft}
\end{minipage}
 &
\begin{minipage}[t]{0.15\textwidth}
\begin{flushleft}\hspace{-0.075in}\fbox{4}\end{flushleft}
\end{minipage}
 &
\begin{minipage}[t]{0.15\textwidth}
\begin{flushleft}\hspace{-0.075in}\fbox{5}\end{flushleft}
\end{minipage}
\\
\includegraphics[width=0.15\textwidth]{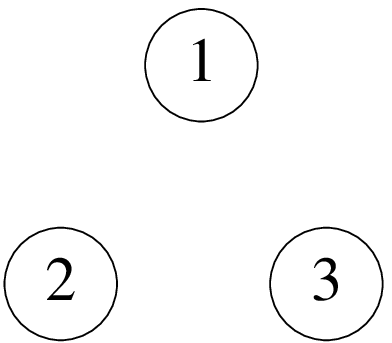}
 & \includegraphics[width=0.15\textwidth]{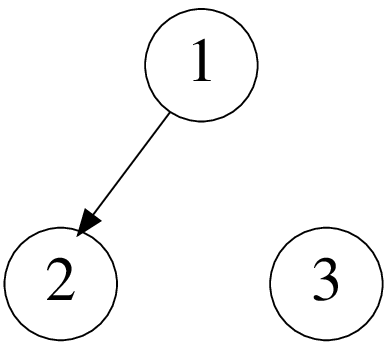} &
\includegraphics[width=0.15\textwidth]{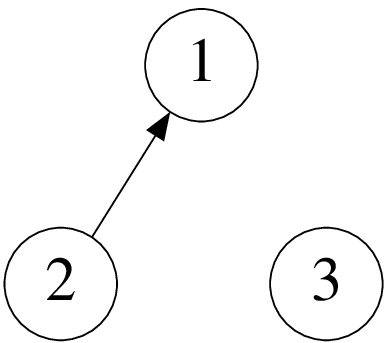} &
\includegraphics[width=0.15\textwidth]{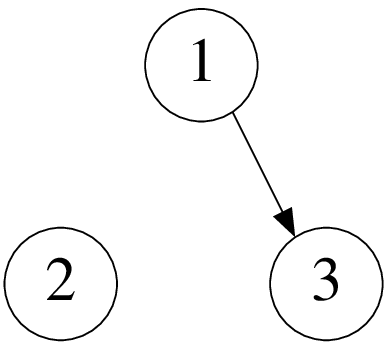} &
\includegraphics[width=0.15\textwidth]{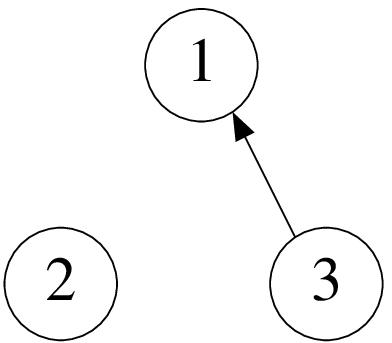} \\
\hline \hline
\begin{minipage}[t]{0.15\textwidth}
\begin{flushleft}\hspace{-0.075in}\fbox{6}\end{flushleft}
\end{minipage}
 &
\begin{minipage}[t]{0.15\textwidth}
\begin{flushleft}\hspace{-0.075in}\fbox{7}\end{flushleft}
\end{minipage}
 &
\begin{minipage}[t]{0.15\textwidth}
\begin{flushleft}\hspace{-0.075in}\fbox{8}\end{flushleft}
\end{minipage}
 &
\begin{minipage}[t]{0.15\textwidth}
\begin{flushleft}\hspace{-0.075in}\fbox{9}\end{flushleft}
\end{minipage}
 &
\begin{minipage}[t]{0.15\textwidth}
\begin{flushleft}\hspace{-0.075in}\fbox{10}\end{flushleft}
\end{minipage}
\\
\includegraphics[width=0.15\textwidth]{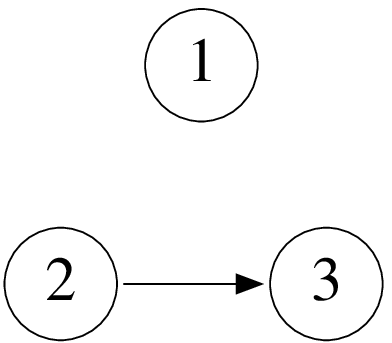}
 & \includegraphics[width=0.15\textwidth]{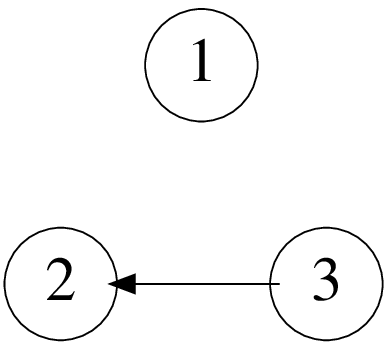} &
\includegraphics[width=0.15\textwidth]{3nodes8.eps} &
\includegraphics[width=0.15\textwidth]{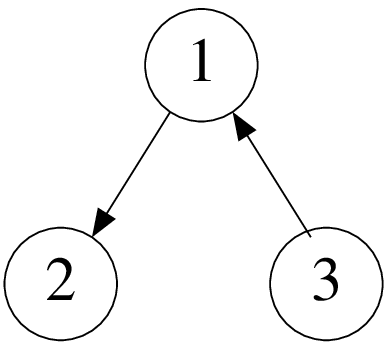} &
\includegraphics[width=0.15\textwidth]{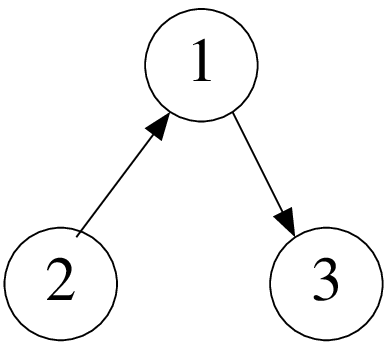} \\ \hline
\begin{minipage}[t]{0.15\textwidth}
\begin{flushleft}\hspace{-0.075in}\fbox{11}\end{flushleft}
\end{minipage}
 &
\begin{minipage}[t]{0.15\textwidth}
\begin{flushleft}\hspace{-0.075in}\fbox{12}\end{flushleft}
\end{minipage}
 &
\begin{minipage}[t]{0.15\textwidth}
\begin{flushleft}\hspace{-0.075in}\fbox{13}\end{flushleft}
\end{minipage}
 &
\begin{minipage}[t]{0.15\textwidth}
\begin{flushleft}\hspace{-0.075in}\fbox{14}\end{flushleft}
\end{minipage}
 &
\begin{minipage}[t]{0.15\textwidth}
\begin{flushleft}\hspace{-0.075in}\fbox{15}\end{flushleft}
\end{minipage}
\\
\includegraphics[width=0.15\textwidth]{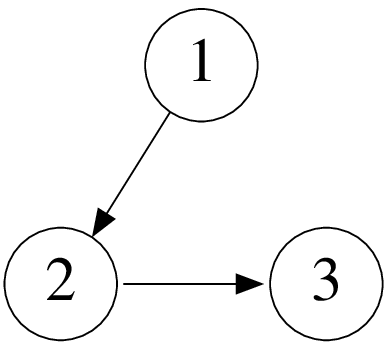}
 & \includegraphics[width=0.15\textwidth]{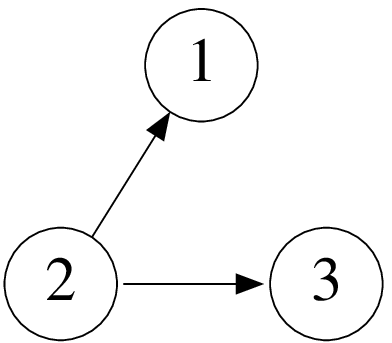} &
\includegraphics[width=0.15\textwidth]{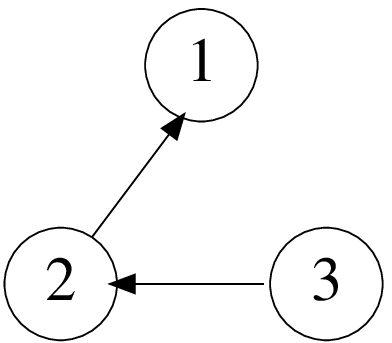} &
\includegraphics[width=0.15\textwidth]{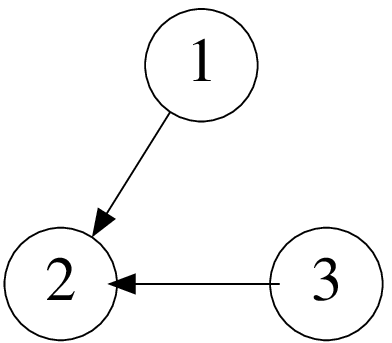} &
\includegraphics[width=0.15\textwidth]{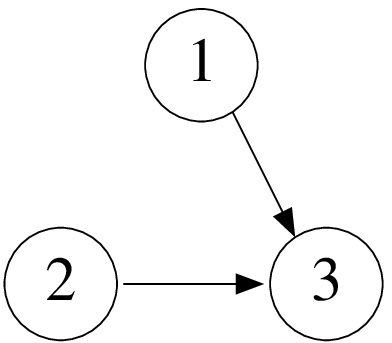} \\ \hline
\begin{minipage}[t]{0.15\textwidth}
\begin{flushleft}\hspace{-0.075in}\fbox{16}\end{flushleft}
\end{minipage}
 &
\begin{minipage}[t]{0.15\textwidth}
\begin{flushleft}\hspace{-0.075in}\fbox{17}\end{flushleft}
\end{minipage}
 &
\begin{minipage}[t]{0.15\textwidth}
\begin{flushleft}\hspace{-0.075in}\fbox{18}\end{flushleft}
\end{minipage}
 &
\begin{minipage}[t]{0.15\textwidth}
\begin{flushleft}\hspace{-0.075in}\fbox{19}\end{flushleft}
\end{minipage}
 &
\begin{minipage}[t]{0.15\textwidth}
\begin{flushleft}\hspace{-0.075in}\fbox{20}\end{flushleft}
\end{minipage}
\\
\includegraphics[width=0.15\textwidth]{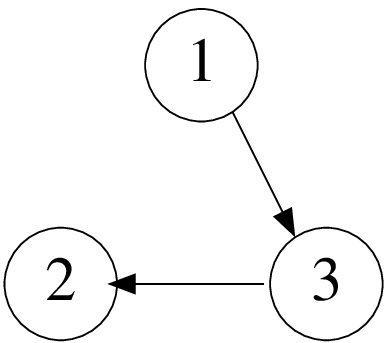}
 & \includegraphics[width=0.15\textwidth]{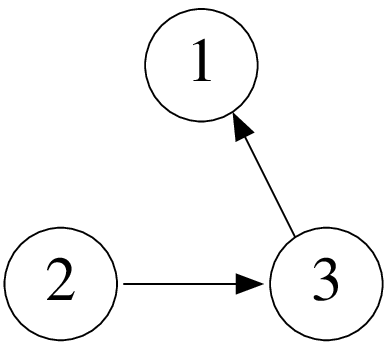} &
\includegraphics[width=0.15\textwidth]{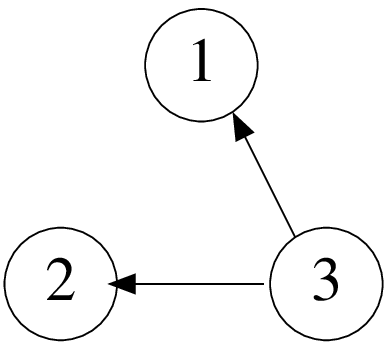} &
\includegraphics[width=0.15\textwidth]{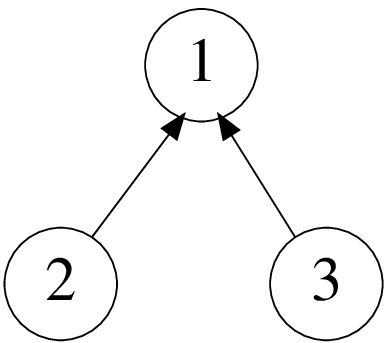} &
\includegraphics[width=0.15\textwidth]{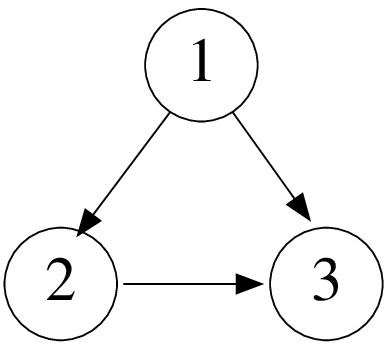} \\ \hline
\begin{minipage}[t]{0.15\textwidth}
\begin{flushleft}\hspace{-0.075in}\fbox{21}\end{flushleft}
\end{minipage}
 &
\begin{minipage}[t]{0.15\textwidth}
\begin{flushleft}\hspace{-0.075in}\fbox{22}\end{flushleft}
\end{minipage}
 &
\begin{minipage}[t]{0.15\textwidth}
\begin{flushleft}\hspace{-0.075in}\fbox{23}\end{flushleft}
\end{minipage}
 &
\begin{minipage}[t]{0.15\textwidth}
\begin{flushleft}\hspace{-0.075in}\fbox{24}\end{flushleft}
\end{minipage}
 &
\begin{minipage}[t]{0.15\textwidth}
\begin{flushleft}\hspace{-0.075in}\fbox{25}\end{flushleft}
\end{minipage}
\\
\includegraphics[width=0.15\textwidth]{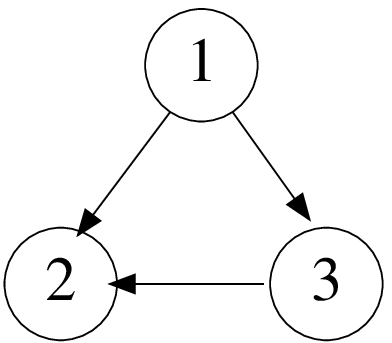}
 & \includegraphics[width=0.15\textwidth]{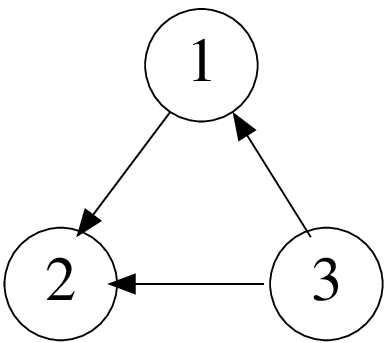} &
\includegraphics[width=0.15\textwidth]{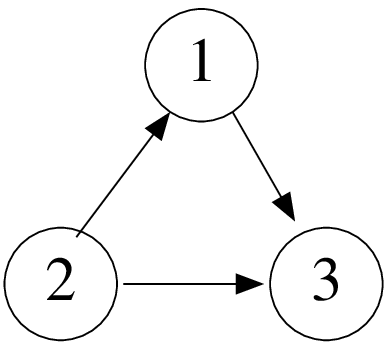} &
\includegraphics[width=0.15\textwidth]{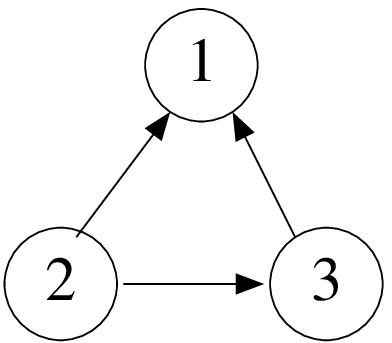} &
\includegraphics[width=0.15\textwidth]{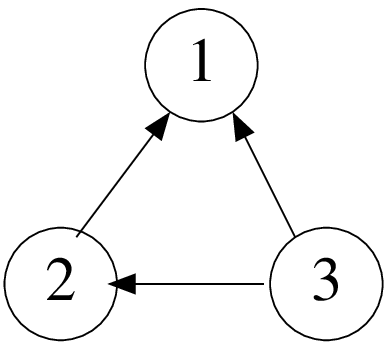} \\ \hline \hline
\end{tabular}
\end{center}
\end{table}

We fixed arbitrary parameters $\theta_{ijk}$ for DAG 8 in Table \ref{t:3nodeDAGs} and simulated $100,000$ values of $(X_{1},X_{2},X_{3})$. (Each $X_{i}$ was assumed to take on values in $\{1,2,\ldots,6\}$.)

Results are shown in Table \ref{t:recovermdl} along with the previously computed values of AIC and BIC. The AIC, BIC, and MDL scores all recovered the correct network ``up to Markov equivalence''.\footnote{A set of DAGs with the same set of edges are said to be \emph{Markov equivalent} if they share the same set of conditional independence relations among variables.} Recovering the specific graph within a Markov equivalence class (a causality problem) requires experimental as opposed to simply observed data and is not the subject of this paper.

\begin{table}[h!]
\caption{AIC, BIC and MDL Recovery}
\label{t:recovermdl}
\begin{center}
\begin{tabular}{|c|c|c|c|}
\hline
Graph & AIC & BIC & MDL\\
Number $(n)$&&&\\
\hline
1& 1040886.085& 1041057.317& 520443.714\\ \hline
2 & 1001501.922 & 1001958.543& 500885.425\\ \hline
3 & 1001501.961 & 1001958.543& 500885.425\\ \hline
4& 991255.256 & 991711.879&495762.093\\ \hline
5& 991255.259 & 991711.879&495762.093\\ \hline
6 & 1036182.434 & 1036639.054&518225.681\\ \hline
7 & 1036182.434 & 1036639.054&518225.681\\ \hline
\LCC
\lightgray & \lightgray & \lightgray &\lightgray\\
8& 951871.096 & 952613.104&476213.804\\ 
\ECC
\hline
9& 951871.096 & 952613.104&476213.804\\ \hline
10& 951871.096 & 952613.104&476213.804\\ \hline
11 & 996798.271& 997540.279&498677.392\\ \hline
12 & 996798.271 & 997540.279&498677.392\\ \hline
13 & 996798.271 & 997540.279&498677.392\\ \hline
14 & 991435.745& 936604.692&496536.904\\ \hline
15 & 986551.608& 987293.616&493554.060\\ \hline
16& 986551.608&987293.616 & 493554.060\\ \hline
17& 986551.608&987293.616 & 493554.060\\ \hline
18& 956755.233&958924.180 & 479196.648\\ \hline
19& 952051.582&954505.917 & 476988.615\\ \hline
20& 952051.582&954505.917 & 476988.615\\ \hline
21& 952051.582&954505.917 & 476988.615\\ \hline
22& 952051.582&954505.917 & 476988.615\\ \hline
23& 952051.582&954505.917 & 476988.615\\ \hline
24& 952051.582&954505.917 & 476988.615\\ \hline
25& 952051.582&954505.917 & 476988.615\\ \hline
\end{tabular}
\end{center}
\end{table}

\section{Minimum Description Length for Discretization}
\slabel{mdldisc}

A discretization of data for the random variable $X_{i}$, represented by node $i$, is a mapping from the range of values in the data set to the set $\{1,2,\ldots, k_{i}\}$ for some $k_{i} \geq 1$. It can be described by ordering the distinct values observed for $X_{i}$ and inserting up to $k_{i}-1$ ``thresholds''. For example

\beq
\elabel{discex}
\underbrace{0.38  \,\,\,\,\,\, 0.42 \,\,\,\,\,\, 0.53 \,\,\,\,\,\, 0.71}_{\mbox{map to 1}} \,\,\,|\,\,\, \underbrace{1.37\,\,\,\,\,\,  1.94 \,\,\,\,\,\ 2.10}_{\mbox{map to 2}} \,\,\,|\,\,\, \underbrace{5.38 \,\,\,\,\,\,  7.11}_{\mbox{map to 3}}.
\eeq

In this paper, we will assume that only node $i$ needs to be discretized and that the remaining nodes are discrete or have already been discretized. We refer the reader to  Friedman and Goldszmidt \cite{friedgold96} for a discussion of multiple node discretization which essentially involves fixing discretizations for all but one node, discretizing that node, and repeating the process by cycling through all nodes repeatedly.

As before, we assume that we have $m$ observations of the $n$-dimensional $(X_{1}, X_{2}, \ldots, X_{n})$. Let $m_{i}$ be the number of distinct values taken on by $X_{i}$ in the data set. Note that $m_{i} \leq m$, with equality possible only for truly continuous data. Define $X_{i}^{*}$ as the discretized version of $X_{i}$. To discretize to  $k_{i} \leq m_{i}$ values, we need to choose $k_{i}-1$ thresholds to put in $m_{i}-1$ spaces between ordered values. There are $\left( \begin{array}{c} m_{i}-1\\k_{i}-1 \end{array} \right)$ threshold configurations to consider. Since we will not know in advance how many values we should have in the discretized data set, we need to consider everything from $k_{i}=1$, which corresponds to mapping all values for $X_{i}$ in the data set to the single value of 1, to $k_{i}=m_{i}$, which corresponds to no discretization at all. In total, there are
$$
\left(
\begin{array}{c}
m_{i}-1\\
0
\end{array}
\right)+
\left(
\begin{array}{c}
m_{i}-1\\
1
\end{array}
\right)
+\ldots
\left(
\begin{array}{c}
m_{i}-1\\
m_{i}-1
\end{array}
\right)=2^{m_{i}-1}
$$
discretizations to consider.

Friedman and Goldszmidt \cite{friedgold96} define a description length score for a network with $X_{i}$ discretized into a particular configuration of $k_{i}$ thresholds using essentially four terms. These terms include $DL_{net}$ and $DL_{data}$, previously described in (\ref{e:DLnet}) and (\ref{e:DLdata}), computed now after discretization-- we will call these terms $DL_{net}^{*}$ and $DL_{data}^{*}$. Also included are terms that encode the index denoting a particular discretization policy and description length for information needed to recover the original data from the discretized data.

\subsection*{Encoding the Discretization Policy}

For fixed $m_{i}$ and $k_{i}$, there are $\left( \begin{array}{c} m_{i}-1\\ k_{i}-1  \end{array} \right)$ different possible configurations for thresholds. Assume we have labeled them from $1$ to $\left( \begin{array}{c} m_{i}-1\\ k_{i}-1  \end{array} \right)$. Storing the index for a particular policy will take at most $\left\lceil \log \left( \begin{array}{c} m_{i}-1\\ k_{i}-1  \end{array} \right) \right\rceil$ bits. Friedman and Goldszmidt use a conservative upper bound based on the inequality
$$
\left( \begin{array}{c} n\\ k \end{array} \right) \leq 2^{nH(k/n)},
$$
where 
$$
H(p) := -p \log p -(1-p) \log (1-p),
$$
and conservatively reserve
\beq
\elabel{DLdp}
DL_{DP} = (m_{i}-1) H \left( \frac{k_{i}-1}{m_{i}-1} \right)
\eeq
to encode the discretization policy. (In practice, we define $H(0)=H(1)=0$.)

\subsection*{Encoding Recovery of Original Data}

Consider again the example of 9 ($m_{i}=9$) distinct values for $X_{i}$ given by (\ref{e:discex}) from a data set with $m \geq m_{i}$ values. Every time we assign a discretized value to an original value, we should store the original value for recovery. Instead, however, we will store the Shannon binary code for the original value using estimated conditional probabilities based on the entire data set. For example, the value $2.10$ might appear in the entire data set one-fourth of the time. That is, $\widehat{P}(X_{i}=2.10)=1/4$. Among the instances of $1.37$, $1.94$, and $2.10$, it might appear half of the time. That is, $\widehat{P}(X_{i}=2.10|X_{i}^{*}=2)=1/2$. Given a discretized value of $X_{i}^{*}=2$, we will encode the original value, $X_{i}$, using approximately $-\log \widehat{P}(X_{i}|X_{i}^{*}=2)$ bits. In total, for recovering original data from discretized data, we add
\beq
\elabel{DLrecover}
DL_{rec} = - \sum_{i=1}^{m} \log \widehat{P}(X_{i}|X_{i}^{*})
\eeq
to the description length.

\divline

In summary, Friedman and Goldszmidt \cite{friedgold96} define the {\emph{description length discretization score}} as
\beq
\elabel{discscore}
DL^{*} = (m_{i}-1) H \left(  \frac{k_{i}-1}{m_{i}-1} \right) + DL_{net}^{*} + DL_{data}^{*} + DL_{rec}.
\eeq

Given a data set and a particular network structure, one scores various discretization mappings for $X_{i}$ and chooses the discretization that minimizes (\ref{e:discscore}). For a given network, the score in (\ref{e:discscore}) will change over discretizations only in terms directly linked to the $i$th node. Thus, as Friedman and Goldszmidt point out, we only need to consider the ``local description length score'' defined, by picking out relevant terms, as
\beq
\elabel{DLlocal}
\begin{array}{lcl}
DL_{local} &=& (m_{i}-1) H \left( \frac{k_{i}-1}{m_{i}-1} \right) + \log k_{i} \\
&&+ \frac{1}{2} \log m \left[ ||\Pi_{i}|| (k_{i}-1) + \sum_{j: X_{i} \in \Pi_{j}} ||\Pi_{j}^{*}||(||X_{j}||-1)\right]\\
&& -m \left[ \widehat{I}(X_{i}^{*}, \Pi_{i}) + \sum_{j: X_{i} \in \Pi_{j}} \widehat{I}(X_{j}, \Pi_{j}^{*}) \right].
\end{array}
\eeq
Here, $\Pi_{j}^{*}$ is a the set of parents for node $j$, denoted with an asterisk since it  includes the discretized $X_{i}^{*}$, and, 
\beq 
\elabel{mi}
\widehat{I}(\vec{X},\vec{Y}) = \sum_{\vec{x},\vec{y}} \widehat{P}(\vec{X}=\vec{x},\vec{Y}=\vec{y}) \cdot \log \left( \frac{\widehat{P}(\vec{X}=\vec{x}, \vec{Y}=\vec{y})}{\widehat{P}(\vec{X}=\vec{x}) \widehat{P}(\vec{Y}=\vec{y})} \right)
\eeq
is the estimated mutual information between random vectors $\vec{X}$ and $\vec{Y}$.  ($\widehat{I}(\vec{X},\emptyset)$ is defined to be zero.)

\section{Minimizing the Local Description Length Score}
\slabel{minimize}
Computation of a single value of (\ref{e:DLlocal}), which is based on network structure through the information terms, can be quite time consuming. For even moderately sized $m_{i}$, computation of (\ref{e:DLlocal}) repeatedly to check all $2^{m_{i}-1}$ discretization policies can be prohibitive, and when multiple/all nodes need to be discretized (a procedure that involves several passes through each node as discussed in \cite{friedgold96}), computation of (\ref{e:DLlocal}) becomes almost impossible. Friedman and Goldszmidt \cite{friedgold96} give some further computational simplifications and suggest a greedy search routine.

The purpose of this paper is to provide an alternative efficient search strategy for the smallest $DL_{local}$ score. We assert that, for a single node discretization, one need only check $m_{i}$ values of $DL_{local}$ as opposed to $2^{m_{i}-1}$. Multiple continuous nodes can then be cycled for discretization just as Friedman and Goldszmidt have suggested. 

For notational simplicity, we assume, for the remainder of this paper, that the node to be discretized is labeled as node $1$. Also, as we are comparing values of $DL_{local}$ for various discretizations of $X_{1}$, we will drop all asterisk superscript notation, as it is understood that we are considering discretized values.

\subsection*{Single Threshold Top-Down Search Strategy}

Let $DL_{local}(0)$ be (\ref{e:DLlocal}) with all $m_{1}-1$ thresholds in place. For $j=1,2, \ldots, m_{1}-1$, let $DL_{local}(-j)$ be (\ref{e:DLlocal}) with all thresholds except for the $j$th threshold. 

\begin{center}
\fbox{
\begin{minipage}[t]{5.8in}
In order to minimize $DL_{local}$ over all $2^{m_{1}-1}$ discretization policies for  $X_{1}$, make comparisons of $DL_{local}(0)$ with $DL_{local}(-j)$ for $j=1,2,\ldots, m_{1}-1$. If $DL_{local}(-j) \leq DL_{local}(0)$, remove  the $j$th threshold.  
\end{minipage}
}
\end{center}

We first consider the effectiveness of this search strategy in the very ideal situation where we augment given discrete data by introducing superfluous values for node 1 in a larger discrete set. For example, we might replace values of $5$ in the original data set with values in $\{5,6,7\}$ with some arbitrary probabilities, whereupon we hope that our search strategy will minimize $DL_{local}$ and that the configuration of thresholds that does such will correctly map values in $\{5,6,7\}$ back to the original value of $5$. We will prove, in this case, that the ``single threshold top-down'' search strategy will find the discretization policy that minimizes $DL_{local}$ among all $2^{m_{1}-1}$ discretizations for a large enough sample size $m$.

 Indeed, $m$, $m_{1}$, $||\Pi_{1}||$, and $||X_{j}||$ are constant, and $0 \leq H(p) \leq 1$. 
Note that
\beq
\elabel{firstpart}
(m_{1}-1) H \left( \frac{k_{1}-1}{m_{1}-1} \right) + \log k_{1} 
+ \frac{1}{2} \log m \left[ ||\Pi_{1}|| (k_{1}-1) + \sum_{j: X_{1} \in 
\Pi_{j}} ||\Pi_{j}||(||X_{j}||-1)\right]
\eeq
will be constant over any discretization policy for $X_{1}$ with a fixed number of thresholds. In particular, when comparing all possible single threshold removals, starting with any fixed number of thresholds, we can restrict our attention to maximizing
\beq
\elabel{secondpart}
\widehat{I}(X_{1}, \Pi_{1}) + \sum_{j: X_{1} \in \Pi_{j}} \widehat{I}(X_{j}, \Pi_{j}).
\eeq
In the next Section, we consider what our search strategy does to (\ref{e:secondpart}) in the absence of (\ref{e:firstpart}) in an ideal situation where there is a ``correct" discretization. We will see that it maximizes (\ref{e:secondpart}) and that it does so while leaving a minimal number of thresholds. In \Section{closer1}, we will consider (\ref{e:firstpart}) and conclude that we are indeed minimizing $DL_{local}$. We will also see that it finds the correct discretization.

\subsection{A Closer look at Information Terms in an Ideal Situation}
\slabel{closer2}
As the notation to follow gets a bit cumbersome in the general case, we illustrate the proof of most claims in this Section and the next with a concrete example. Consider a two-node network where node $1$ is a parent to node $2$, and assume that that both nodes take on values in $\{1,2,3\}$. From $m$ data points, we can produce estimates 
$$
\begin{array}{lcl}
\widehat{p}(i,j) &:=& \widehat{P}(X_{1}=i,X_{2}=j)\\
\widehat{p}_{1}(i)&:=&\widehat{P}(X_{1}=i), \,\,\,  \mbox{and}\\
\widehat{p}_{2}(j)&:=&\widehat{P}(X_{2}=j),
\end{array}
$$ 
for $i,j, \in \{1,2,3\}$.

In general, for this two node network, we would assume that node 1 takes on values in $\{1,2,\ldots,m_{1}\}$, node 2 takes on values in $\{1,2,\ldots, m_{2}\}$, and one would work with the estimates $\widehat{p}(i,j)$, $\widehat{p}_{1}(i)$, and $\widehat{p}_{2}(j)$
for $i \in \{1,2, \ldots, m_{1}\}$ and $j \in \{1,2,\ldots, m_{2}\}$.

We now ``explode" the data for our specific example  at node $1$ into values in $\{1,2,3,4,5,6\}$ by 
replacing instances of $1$ with values in $\{1,2\}$ with probabilities $1/3$ and $2/3$, respectively, replacing original instances of $2$ with values in $\{3,4,5\}$ with probabilities $2/7$, $4/7$, and $1/7$, respectively, and replacing original instances of $3$ with the value $6$.  Thus, in our ``exploded data set'', node $1$ is taking on values in $\{1,2,3,4,5,6\}$ with probabilities denoted as $\widetilde{p}_{1}(i)$ for $i=1,2,3,4,5,6$, where, for example, $\widetilde{p}_{1}(1) = \frac{1}{3} \widehat{p}_{1}(1)$ and $\widetilde{p}_{1}(4) = \frac{4}{7} \widehat{p}_{1}(2)$. Joint probabilities for $X_{1}$ and $X_{2}$ are denoted by $\widetilde{p}(i,j)$ where we have, for example,
$$
\widetilde{p}(1,j) = \frac{1}{3}\widehat{p}(1,j) \qquad \mbox{and} \qquad 
\widetilde{p}(4,j) = \frac{4}{7}\widehat{p}(2,j).
$$

In the more general case, we could ``explode'' the data at node $1$ more generally by replacing  instances of $1$ with values in $\{1,2, \ldots, \ell_{1}\}$ some probabilities $q(1,1), q(1,2), \ldots, q(1,\ell_{1})$, summing to 1, replacing original instances of $2$ with values in $\{\ell_{1}+1,\ell_{1}+2, \ldots, \ell_{1}+\ell_{2} \}$ with respective probabilities $q(2,1), q(2,2), \ldots, q(2,\ell_{2})$, summing to 1, and so forth.

By ``exploding" the data at node one, we have introduced superfluous values for $X_{1}$ in terms of probabilities for $X_{2}$. For example, $\widetilde{P}(X_{2}=j|X_{1}=1) = \widetilde{P}(X_{2}=j|X_{1}=2)$ for all $j \in \{1,2,3\}$. Any proper discretization process should aggregate the values $1$ and $2$ for $X_{1}$ back into one value. Here, $\widetilde{P}(X_{1}=i,X_{2}=j)$ is used to denote the probability $\widetilde{p}(i,j)$. Similarly, $\widetilde{p}_{1}(i)$ may be denoted as $\widetilde{P}(X_{1}=i)$.

For this two-node network, (\ref{e:secondpart}) is simply $\widehat{I}(X_{1},X_{2})$, which we will denote by $\widehat{I}$.
Define this estimated information as
\beq
\elabel{estinf}
\widehat{I} := \sum_{i,j} \widehat{I}_{i,j} 
\eeq
where 
\beq 
\elabel{estinfterm}
\widehat{I}_{ij}:= \widehat{p}(i,j) \cdot \log \left( \frac{\widehat{p}(i,j)}{\widehat{p}_{1}(i) \cdot \widehat{p}_{2}(j)} \right)
\eeq
and the sums run over $i \in \{1,2,3\}$ and $j \in \{1,2,3\}$.  

Define the corresponding information $\widetilde{I}$ and information terms $\widetilde{I}_{i,j}$  using $\widetilde{p}(i,j)$ in place of $\widehat{p}(i,j)$ with sums running over $i \in \{1,2,\ldots, 6\}$ and $j \in \{1,2,3\}$.

It is easy to verify information term relationships such as
$$
\widetilde{I}_{1,j} = \frac{1}{3}  \widehat{I}_{1,j} \qquad \mbox{and} \qquad \widetilde{I}_{4,j} = \frac{4}{7} \cdot \widehat{I}_{2,j}, 
$$
and consequently that 
$$
\widetilde{I} = \sum_{\tiny \begin{array}{c} i \in \{1,2, \ldots 6 \}\\ j \in \{1,2,3\}\end{array}} \widetilde{I}_{ij} =  \sum_{\tiny \begin{array}{c} i \in \{1,2,3 \}\\ j \in \{1,2,3\}\end{array}} \widehat{I}_{ij} = \widehat{I}.
$$ 
That is, we did not change the information between $X_{1}$ and $X_{2}$ by exploding the data at $X_{1}$. 

We now show that our single threshold top-down search strategy will correctly recover the original values for $X_{1}$. We call this the ``correct discretization'', denoted as $12|345|6$, which means that we will remove three thresholds from the ``full discretization'', denoted as $1|2|3|4|5|6$, and that the values $1$ and $2$ will map back to $1$, the values in $\{3,4,5\}$ will map back to $2$, and the value $6$, will map back to $3$. We will assume that the original values in $\{1,2,3\}$ are all distinct in the sense that $\widehat{P}(X_{2}=j|X_{1}=1) \ne \widehat{P}(X_{2}=j|X_{1}=2)$ for some $j$, $\widehat{P}(X_{2}=j|X_{1}=1) \ne \widehat{P}(X_{2}=j|X_{1}=3)$ for some $j$, and $\widehat{P}(X_{2}=j|X_{1}=2) \ne \widehat{P}(X_{2}=j|X_{1}=3)$ for some $j$, so that they should not be aggregated further.

\subsection*{Removing a Threshold: The Impact on Information}
Starting with the exploded data, with values for $X_{1}$ represented as $1|2|3|4|5|6$, we consider the $(X_{1},X_{2})$ information term after removing the threshold between the values $r$ and $r+1$ for some $r \in \{1,2,\ldots,5\}$. Note that removal of this $r$th threshold will leave all values below the threshold unchanged, while all values above will be decreased by 1. For example, if we remove the third threshold, we denote the new configuration as $1|2|34|5|6$, but it represents a mapping
$$
\underbrace{1}_{\mbox{map to 1}} \,\,\,|\,\,\, \underbrace{2}_{\mbox{map to 2}} \,\,\,|\,\,\, \underbrace{3 \,\,\,  4}_{\mbox{map to 3}}\,\,\,|\,\,\, \underbrace{5}_{\mbox{map to 4}} \,\,\,|\,\,\, \underbrace{6}_{\mbox{map to 5}}.
$$

\vspace{0.2in}
Define, for $i \in \{1,2,\ldots, 5\}$ and $j \in \{1,2,3\}$, the joint and marginal probabilities for $X_{1}$ and $X_{2}$ after the $r$th threshold is removed as $p^{(r)}(i,j)$, $p^{(r)}_{1}(i)$, and $p^{(r)}_{2}(j)$. We have, for $j \in \{1,2,3\}$, the relationships 
$$
\begin{array}{lcl}
p^{(r)}(i,j) &=& \widetilde{p}(i,j), \qquad  i \in \{1,2, \ldots, r-1\} \,\,\, (r>1)\\
p^{(r)}(r,j) &=& \widetilde{p}(r,j) + \widetilde{p}(r+1,j)\\
p^{(r)}(i,j) &=& \widetilde{p}(i+1,j), \qquad i \in \{r+1, r+2, \ldots, 6 \}, \,\, (r < 5)\\
\\
p^{(r)}_{1}(i) &=& \widetilde{p}_{1}(i), \qquad  i \in \{1,2, \ldots, r-1\}\\
p^{(r)}_{1} (r) &=& \widetilde{p}_{1}(r) + \widetilde{p}_{1}(r+1)\\
p^{(r)}_{1}(i) &=& \widetilde{p}_{1}(i+1), \qquad i \in \{r+1,r+3,\ldots, 6\}, \,\, (r < 5),\\
\end{array}
$$
and
$$
p^{(r)}_{2}(j) = \widetilde{p}_{2}(j) = \widehat{p}_{2}(j).
$$


Defining $I^{(r)}$ and $I^{(r)}_{ij}$ analogous to (\ref{e:estinf}) and (\ref{e:estinfterm}), using $p^{(r)}(i,j)$, we have, for $j \in \{1,2,3\}$,
$$
\begin{array}{lcl}
I_{rj}^{(r)} &=& p^{(r)}(r,j) \cdot \log \left( \frac{p^{(r)}(r,j)}{p^{(r)}_{1}(r) p^{(r)}_{2}(j)} \right)\\
\\
&=& [\widetilde{p}(r,j) + \widetilde{p}(r+1,j)] \cdot \log \left( \frac{\widetilde{p}(r,j) + \widetilde{p}(r+1,j)}{[\widetilde{p}_{1}(r) + \widetilde{p}_{1}(r+1)] \cdot \widetilde{p}_{2}(j)}\right)\\
\\
&\leq& \widetilde{p}(r,j)  \cdot \log \left( \frac{\widetilde{p}(r,j)}{\widetilde{p}_{1}(r)  \cdot \widetilde{p}_{2}(j)}\right) + \widetilde{p}(r+1,j)] \cdot \log \left( \frac{\widetilde{p}(r+1,j)}{ \widetilde{p}_{1}(r+1) \cdot \widetilde{p}_{2}(j)}\right)\\
\\
&=& \widetilde{I}_{rj} + \widetilde{I}_{r+1,j}.
\end{array}
$$
The inequality is  is due to the log-sum inequality,
$$
\sum_{i=1}^{n} a_{i} \log \left(\frac{a_{i}}{b_{i}} \right) \geq \left[\sum_{i} a_{i}\right] \log \left(\frac{\sum_{i} a_{i}}{\sum_{i} b_{i}}\right),
$$ 
which holds for any nonnegative $a_{1}, a_{2}, \ldots, a_{n}$ and $b_{1}, b_{2}, \ldots, b_{n}$. The log-sum inequality can be shown to be an equality if and only if the $a_{i}/b_{i}$ are equal for all $i=1,2,\ldots, n$. Thus, we have that
\beq
\elabel{rinequ}
I_{rj}^{(r)} \leq \widetilde{I}_{rj} + \widetilde{I}_{r+1,j}
\eeq
for $j \in \{1,2,3\}$, with equality if and only if
$$
\frac{\widetilde{p}(r,j)}{\widetilde{p}_{1}(r)  \cdot \widetilde{p}_{2}(j)} = \frac{\widetilde{p}(r+1,j)}{ \widetilde{p}_{1}(r+1) \cdot \widetilde{p}_{2}(j)}.
$$
This  happens if and only if
\beq
\elabel{shoulddisc}
\widetilde{P}(X_{2}=j|X_{1}=r) = \widetilde{P}(X_{2}=j|X_{1}=r+1).
\eeq
for $j \in \{1,2,3\}$, which is precisely when the values $r$ and $r+1$ in the exploded version of $X_{1}$ should be aggregated or discretized into one value. If we do not have (\ref{e:shoulddisc}), aggregating the values will result in a loss of information.

\begin{figure}[htp]
\caption{Correct Removal of a Threshold Corresponds to an Alternate Explosion}
\label{fig:diffexpl}
\begin{center}
\ \psfig{figure=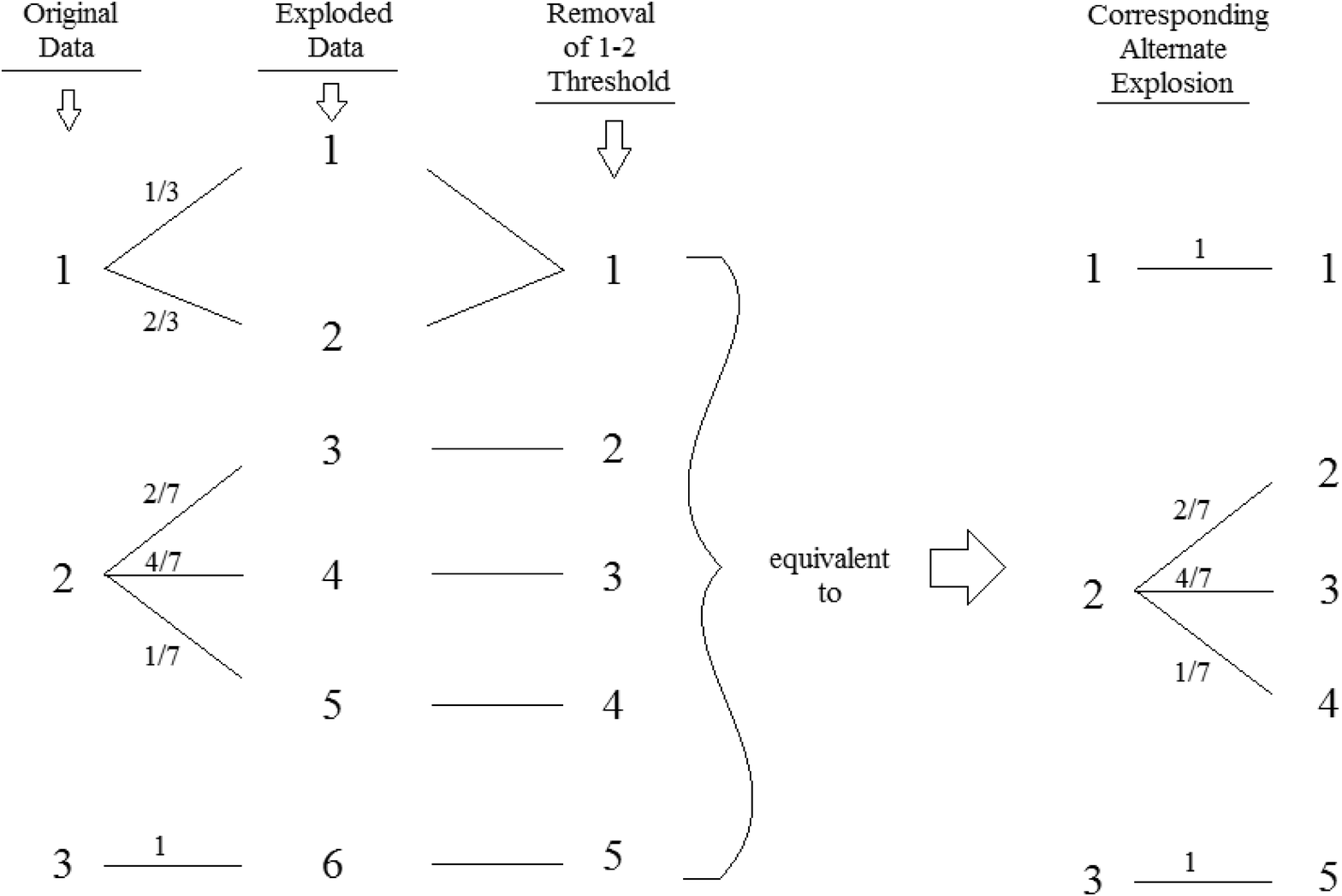,width=\textwidth} \\
\end{center}
\end{figure}

Note that $I^{(r)}$ denotes the value of the information between $X_{1}$ and $X_{2}$ with the $r$th threshold in the explosion for $X_{1}$ removed. 
In our example, 
$$
\widetilde{P}(X_{2}=j|X_{1}=1) = \frac{\widetilde{p}(1,j)}{\widetilde{p}_{1}(1)} = \frac{(1/3)\widehat{p}(1,j)}{(1/3)\widehat{p}_{1}(1)} =\widehat{P}(X_{2}=j|X_{1}=1)
$$
and
$$
\widetilde{P}(X_{2}=j|X_{1}=2) = \frac{\widetilde{p}(2,j)}{\widetilde{p}_{2}(1)} = \frac{(2/3)\widehat{p}(1,j)}{(2/3)\widehat{p}_{1}(1)} =\widehat{P}(X_{2}=j|X_{1}=1).
$$
Thus, we have (\ref{e:shoulddisc}), when $r=1$, and consequently  
$$
\begin{array}{lcl}
I^{(1)} &=&    \sum_{i=1}^{5}\sum_{j=1}^{3} I^{(1)}_{ij}\\
\\
&=& \sum_{j=1}^{3} I^{(1)}_{1j}  + \sum_{i=2}^{5}\sum_{j=1}^{3} I^{(1)}_{ij}\\
\\
&=&   \sum_{j=1}^{3} I^{(1)}_{1j}  + \sum_{i=3}^{6} \sum_{j=1}^{3}\widetilde{I}_{ij}\\
\\
&=& \sum_{j=1}^{3} (\widetilde{I}_{1j} + \widetilde{I}_{2j})  + \sum_{i=3}^{6}\sum_{j=1}^{3} \widetilde{I}_{ij}\\
\\
&=& \sum_{i=1}^{6} \widetilde{I}_{ij} = \widetilde{I}.
\end{array}
$$
On the other hand, considering the second threshold removal from the full discretization $1|2|3|4|5|6$, similar calculations show that
$$
\widetilde{P}(X_{2}=j|X_{1}=2) = \widehat{P}(X_{2}=j|X_{1}=1)
$$
which is not, in general, equal to
$$
\widetilde{P}(X_{2}=j|X_{1}=3) = \widehat{P}(X_{2}=j|X_{1}=2),
$$
so we have the strict inequality $I^{(2)}_{2,j} < \widetilde{I}_{2,j} + \widetilde{I}_{3j}$ and therefore
$$
\begin{array}{lcl}
I^{(2)} &=&    \sum_{i=1}^{5}\sum_{j=1}^{3} I^{(2)}_{ij}\\
\\
&=& \sum_{j=1}^{3} I^{(2)}_{1j} + \sum_{j=1}^{3} I^{(2)}_{2j}  + \sum_{i=3}^{5}\sum_{j=1}^{3} I^{(2)}_{ij}\\
\\
&=& \sum_{j=1}^{3} \widetilde{I}_{1j} + \sum_{j=1}^{3} I^{(2)}_{2j}  + \sum_{i=4}^{6}\sum_{j=1}^{3} \widetilde{I}_{ij}\\
\\
&< & \sum_{j=1}^{3} \widetilde{I}_{1j} + \sum_{j=1}^{3} (\widetilde{I}_{2j}+\widetilde{I}_{3j})  + \sum_{i=4}^{6}\sum_{j=1}^{3} \widetilde{I}_{ij}\\
\\
&=& \sum_{i=1}^{6} \widetilde{I}_{ij} = \widetilde{I}.
\end{array}
$$
In all, we can show in this way that
\beq
\elabel{results}
I^{(1)} = \widetilde{I}, \,\,\, I^{(2)} < \widetilde{I}, \,\,\, I^{(3)} = \widetilde{I}, \,\,\, I^{(4)} = \widetilde{I}, \,\,\,\mbox{and} \,\,\, I^{(5)} < \widetilde{I}.
\eeq
So, our search strategy will produce the correct discretization $12|345|6$. The question remains though as to whether this is actually the discretization that minimizes $DL_{local}$. Due to (\ref{e:rinequ}), we will always have $I^{(r)} \leq \tilde{I}$. In fact, due to the log-sum inequality, any removal of a threshold from any configuration with any number of thresholds can never increase information. Thus, the full discretization $1|2|3|4|5|6$ will always have maximal information. From (\ref{e:results}), we see that the discretizations $12|3|4|5|6$, $1|2|34|5|6$, and $1|2|3|45|6$ have the same, and thus also maximal, information. Note that each of these three configurations corresponds to a different explosion of the original data. This is illustrated for $12|3|4|5|6$ in  Figure \ref{fig:diffexpl}.
Thus, by the same information arguments above, ``correct removal" (removal of one of the superfluous thresholds) of a threshold from $12|3|4|5|6$ will result in another configuration with the same maximal information. Since this is true starting with any one of the three correct single threshold removal configurations, we see that we can remove two of the superfluous thresholds from the full explosion and still maintain the maximal information. Continuing this argument, we can remove all  superfluous thresholds from the full discretization and the resulting configurations of thresholds, which in this example is $12|345|6$, will have the maximal information.

\subsection{A Closer look at the Leading Terms in  $DL_{local}$}
\slabel{closer1}

For a general DAG, the leading terms in $DL_{local}$, given in (\ref{e:firstpart}) can be rewritten as
$$
(m_{1}-1) H \left(  \frac{k_{1}-1}{m_{1}-1}  \right) + \log k_{1} + \frac{1}{2} (\log m) (c k_{1}- ||\Pi_{1}||) =: D(k_{1})
$$
for some $c \geq 0$ where $c=0$ if and only if node $1$ is not connected to any other nodes. 

The second and third terms here are clearly increasing in $k_{1}$. The first term, as a function of $k_{1}$, is symmetric about $k_{1}=(m_{1}+1)/2$, increasing to the left of this value and decreasing to the right. However, since
$$
\left. \frac{d}{dx} D(x) \right|_{x=k_{1}} = \log \left(  \frac{m_{1}-k_{1}}{k_{1}-1}\right)  + \frac{1}{k_{1}} + \frac{1}{2} \,  c  \cdot \log  m,
$$
we can, when $c>0$ choose $m$ (the sample size) large enough to ensure that $D(k_{1})$ is increasing in $k_{1}$. In this case, this first part of $DL_{local}$ acts as a penalty term and $DL_{local}$ is minimized by choosing the discretization that maximizes the information terms with the minimum number of thresholds. By design, the single threshold top-down search strategy will do exactly this. Thus, when employing the strategy for large samples,  we can  ignore these leading terms.

\subsection{More Complicated Networks}
\slabel{morecomplicated}

The observations in \Section{closer1} were not dependent on the specific network, however, our analysis of the information terms was for a two-node network where $X_{1}$ was a parent to $X_{2}$. Since $\widehat{I}(X_{1},X_{2}) = \widehat{I}(X_{2},X_{1})$, our search strategy will also find the ``optimal  discretization" for $X_{1}$, i.e. the one that  that maximizes (\ref{e:secondpart}) with a minimum number of thresholds, for the two-node network where $X_{1}$ is a child of $X_{2}$.

We will now check that the  strategy will find the optimal discretization for $X_{1}$  for general networks.  To this end, we begin by independently considering the two types of terms in (\ref{e:secondpart}). 

\begin{itemize}
\item $\widehat{I}(X_{1}, \Pi_{1})$

Since replacing $X_{2}$  with a vector of random variables has no effect on any of the computations in \Section{closer2}, we see that our search strategy will find the discretization for $X_{1}$ that maximizes $\widehat{I}(X_{1},\Pi_{1})$ with a minimum number of thresholds.

\item $\widehat{I}(X_{j}, \Pi_{j})$ where $X_{1} \in \Pi_{j}$

Suppose, 
 for ease of exposition, that $j=2$ and, that all nodes originally take values in $\{1,2,3\}$, and that $X_{1}$  has been exploded to take values in $\{1,2,3,4,5,6\}$ just as in, and using the same probabilities as, \Section{closer2}. If $\Pi_{2}$ consists only of $X_{1}$, we have already seen that our search strategy will maximize the information term with a minimum number of thresholds. Assuming now that
$\Pi_{2}$ consists of $X_{1}$ and some vector $\vec{Y}$ whose components are the other parents of $X_{2}$,
we have
$$
\widehat{I}:=\widehat{I}(X_{2},\Pi_{2})  = \widehat{I}(X_{2},(X_{1},\vec{Y})) = \sum_{ij\vec{k}} \widehat{I}_{ij\vec{k}}
$$
where
$$
\widehat{I}_{ij\vec{k}} = \widehat{p}(i,j,\vec{k}) \cdot \log \left( \frac{\widehat{p}(i,j,\vec{k})}{\widehat{p}_{2}(j)  \, \widehat{p}_{1Y}(i,\vec{k})}\right).
$$
Here, $\widehat{p}(i,j,\vec{k}) = \widehat{P}(X_{1}=i,X_{2}=j,\vec{Y}=\vec{k})$, $\widehat{p}_{2}(j) = \widehat{P}(X_{2}=j)$, and $\widehat{p}_{1Y}(i,\vec{k}) = \widehat{P}(X_{1}=i,\vec{Y}=\vec{k})$. Using $\widetilde{p}$ to denote probabilities after the explosion of $X_{1}$, it is easy to see expected relationships such as
$$
\widetilde{p}(4,j,\vec{k}) = \frac{4}{7} \widehat{p}(2,j,\vec{k}), \qquad \widetilde{p}_{2}(j) = \widehat{p}_{2}(j), \qquad \mbox{and} \,\,\,\,\,\, \widetilde{p}_{1Y}(2,\vec{k})=\frac{2}{3} \widehat{p}_{1Y}(1,\vec{k}).
$$
Therefore, we can verify, for example,  that $\widetilde{I}_{2j\vec{k}} = (2/3) \widehat{I}_{1j\vec{k}}$, and that the overall information terms $\widetilde{I}$ and $\widehat{I}$ are equal. We can also verify, using the log-sum inequality, that  after removal of the $r$th threshold we have
$$
I^{(r)}_{r,j,\vec{k}} \leq \widetilde{I}_{r,j,\vec{k}} + \widetilde{I}_{r+1,j,\vec{k}},
$$
with equality if and only if 
\beq
\elabel{3wayinfoterm}
\widetilde{P}(X_{2}=j|X_{1}=r,\vec{Y}=\vec{k}) = \widetilde{P}(X_{2}=j|X_{1}=r+1,\vec{Y}=\vec{k}).
\eeq
Summing over all appropriate values for indices, we get that $I^{(r)}$, the information term between $X_{2}$ and its parents after removal of the $r$th threshold for $X_{1}$, is less than or equal to $\widetilde{I}$, with equality if and only if (\ref{e:3wayinfoterm}) holds for all $j$ and $\vec{k}$. When $r=1$, for example, both sides of (\ref{e:3wayinfoterm}) are equal to $\widehat{P}(X_{2}=j|X_{1}=1, \vec{Y}=\vec{k})$, indicating that the first threshold should be removed. When $r=2$, the left side is equal to $\widehat{P}(X_{2}=j|X_{1}=2, \vec{Y}=\vec{k})$ and the right side is equal to $\widehat{P}(X_{2}=j|X_{1}=3, \vec{Y}=\vec{k})$, so we do not have  (\ref{e:3wayinfoterm}) and hence a decrease in information. Thus, we would not remove the second threshold.

$\widetilde{I}$,  the information with all thresholds in place, is the maximal information among all possible discretizations. If we remove each threshold that leaves the information unchanged, we will still have the maximal information with a minimum number of thresholds. Thus, the top-down search strategy will give the optimal discretization.

\end{itemize}

Considering all terms in (\ref{e:secondpart}) independently may result in different discretizations. For example, consider graph 8 from Table \ref{t:3nodeDAGs}. In this case, (\ref{e:secondpart}) becomes $\widehat{I}(X_{1},X_{2})+\widehat{I}(X_{1},X_{3})$. If $\widehat{P}(X_{2}=j|X_{1}=1) = \widehat{P}(X_{2}=j|X_{1}=2)$ for all $j$ but $\widehat{P}(X_{3}=j|X_{1}=1) \ne \widehat{P}(X_{3}=j|X_{1}=2)$ for some $j$, removing the threshold between $1$ and $2$ for the discretization of $X_{1}$ would leave the information between $X_{1}$ and $X_{2}$ unchanged but would decrease he information between $X_{1}$ and $X_{3}$. However, as both information terms, and hence their sum, are maximized with the full discretization for $X_{1}$, the top down-search strategy will not allow us to remove the threshold between $1$ and $2$. That is, it will only remove thresholds between values of $X_{1}$ that are indistinguishable in terms of the conditional distributions involving all nodes to connected to $X_{1}$.

\section{Conclusions}

We have seen, in the case of ideal ``exploded" data where there is a ``correct" discretization, that the minimum description length scoring mechanism of Friedman and Goldszmidt will in fact recover the discretization. Just as importantly, we have seen that we can find it from among $2^{m_{1}-1}$ possibilities by making only $m_{1}-1$ comparisons. 

In the case of discrete data where superfluous values were not manufactured, for example the two-node network where $X_{1}$ is a parent to $X_{2}$ that originally takes values in $\{1,2,3,4,5,6\}$ and $\{1,2,3\}$, respectively, we should aggregate $1$ and $2$ for node $1$ into a single value if $P(X_{2}=j|X_{1}=1) = P(X_{2}=j|X_{1}=2)$ for all $j \in \{1,2,3\}$. From the data, we will only get to see that $\widehat{P}(X_{2}=j|X_{1}=1) \approx \widehat{P}(X_{2}=j|X_{1}=2)$. Even with a large sample size, because of the approximation, we would still see some decrease in overall information when correctly removing the threshold, so it remains to determine when such a decrease is significant. We have much empirical evidence that we will still be able to recover the correct discretization by comparing $DL_{local}$ for only single threshold removals to $DL_{local}$ for the full discretization. As an example, we simulated $100,000$ values for $X_{1}$ and $X_{2}$ in the two-node network by simulating $X_{1}$ in $\{1,2,3,4,5,6\}$ directly (as opposed to first simulating them in $\{1,2,3\}$ and then exploding the data). We chose parameters such that   $P(X_{2}=j|X_{1}=1) = P(X_{2}=j|X_{1}=2)$ and   $P(X_{2}=j|X_{1}=5) = P(X_{2}=j|X_{1}=6)$ for all $j \in \{1,2,3\}$. In Table \ref{t:dlscore}, we show the full discretization score, all single threshold removal scores, and the true discretization score. The two incorrect threshold removals (between 1 and 2 and between 5 and 6) stand out as having different $DL_{local}$ scores than the rest. The quantification of this difference is still ongoing work.

\begin{table}[h!]
\caption{Local Description Length Score}
\label{t:dlscore}
\begin{center}
\begin{tabular}{|c|c|}
\hline
Discretization & DL local\\
\hline
$1|2|3|4|5|6$	&	-29841.52	\\	\hline
$12|3|4|5|6$	&	-29870.53	\\	\hline
$1|23|4|5|6$	&	-24456.07	\\	\hline
$1|2|34|5|6$	&	-29866.02	\\	\hline
$1|2|3|45|6$	&	-29896.74	\\	\hline
$1|2|3|4|56$	&	-24585.90	\\	\hline \hline \hline
$12|345|6$	&	-29929.17	\\      \hline
\end{tabular}
\end{center}
\end{table}      

Computation of the $DL_{local}$ score requires that one have a graph under consideration. Indeed, all or our arguments in this paper have been made using the correct generating graph. In practice, it seems that we can still recover the correct discretization using {\bf{any}} graph as long as the node to be discretized is connected to at least one other node which, incidentally, is precisely when we can ignore the leading terms in $DL_{local}$ as discussed in \Section{closer1}. We recommend running the search for the discretization that minimized $DL_{local}$ on several networks. Fortunately, our proposed fast search strategy makes this feasible.

The case of truly continuous data is still problematic. For the approach of this paper, it is incorrect to use estimates of densities and information terms for continuous random variables, as the procedures discussed all involve comparisons of scores computed after proposed discretizations. 
Consider again the two node network where $X_{1}$ is a parent of $X_{2}$ and node $1$ is the one to be discretized. (Recall that the MDL discretization process is  for one node at a time. It is assumed that other nodes have already been discretized and that they will be ``re-discretized" pending discretization of the current node in a cyclic fashion.) Relabel the $m_{1}=m$ distinct values observed for  $X_{1}$ as $1$ through $m$. We should aggregate $1$ and $2$ for node $1$ into a single value if $\widehat{P}(X_{2}=j|X_{1}=1) \approx \widehat{P}(X_{2}=j|X_{1}=2)$ for all $j \in \{1,2,3\}$, however approximations of such probabilities will can only be $0$ or $1$! Perhaps then, a bottom-up approach to minimizing $DL_{local}$ would be better though we have not, as yet, had great success with simulated results. This too is ongoing work.



\bibliographystyle{plain}

\bibliography{C:/LaTeX/BIBS/strings,C:/LaTeX/BIBS/masterbib}

\end{document}